# Visual motion analysis of the player's finger

Marco Costanzo

Politecnico di Milano

Milan, Italy

marco.costanzo@mail.polimi.it

23/02/2023

**Abstract**

This work is about the extraction of the motion of fingers (in their three articulations) of a keyboard player from a video sequence.

The relevance of the problem involves several aspects, in fact, the extraction of the movements of the fingers may be used to compute the keystroke efficiency and individual joint contributions, as showed by Werner Goebl and Caroline Palmer in the paper "Temporal Control and Hand Movement Efficiency in Skilled Music Performance" [1].
Those measures are directly related to the precision in timing and force measures.

A very good approach to the hand gesture recognition problem has been presented in the paper " Real-Time Hand Gesture Recognition Using Finger Segmentation" [2] . Another relevant work has been done by  Ali Özkaya and Sidem Işıl Tuncer in their thesis "Pressed Piano Key Detection and Transcription by Visual Motion Analysis"[3].  The authors were able to detect the pressed piano keys from two video sources by analysing the fingers of the piano player.

Detecting the keys pressed on a keyboard is a task that can be complex [4] because of the shadows that can degrade the quality of the result and possibly cause the detection of not pressed keys.  Among the several approaches that already exist, a great amount of them is based on the subtraction of frames in order to detect the movements of the keys caused by their pressure. Detecting the keys that are pressed could be useful to automatically evaluate the performance of a pianist or to automatically write sheet music of the melody that is being played.

# Solution approach

**Extraction of the motion of fingers**

The starting point is two videos taken from two different perspectives of a pianist playing the piano. The videos were taken from above and from the side and in both recordings, the hands of the pianist were captured.

Figure 1. Image taken from above, song1

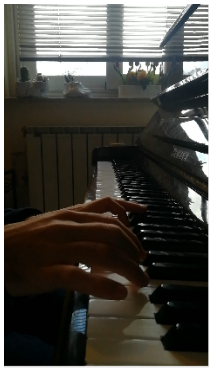

Figure 2. Image taken from the side, song1

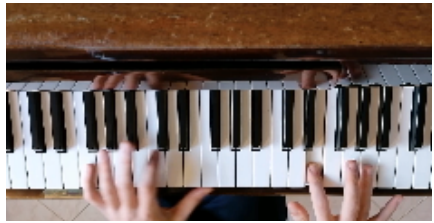

From that point, I proceeded by applying a filter to dehaze those frames.

Figure 3. Image taken from the side (dehaze filter applied), song1

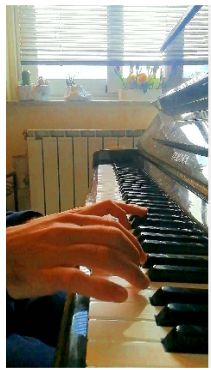

After that, I obtained the frames of the two videos that present a good exposure from which I can proceed to extract the hands of the pianist.

Figure 4. Image taken from the side (dehaze filter applied), song1

Figure 5. Image taken from above (dehaze filter applied), song1

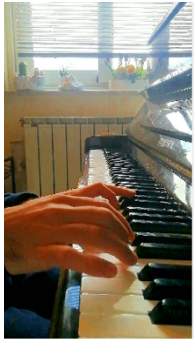
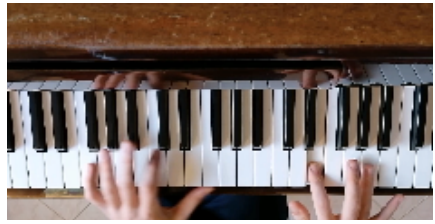

Figure 6. Image taken from the side (dehaze filter applied), song2

Figure 7. Image taken from above(dehaze filter applied), song2

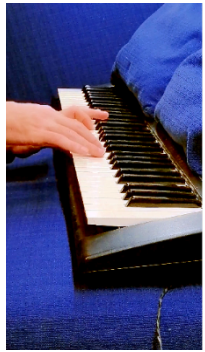
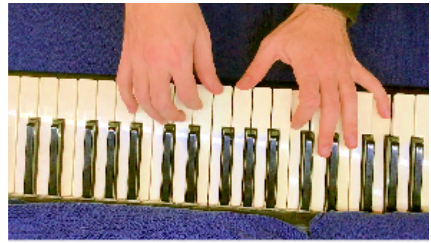

To do so I applied some colour filters that I created in order to filter only the pixels having the red, green and blue components in a suitable range that I identified by sampling some points belonging to the hands of the pianist from the 2 scenes. After an accurate selection of the ranges of the RGB components of the frames, a quite satisfactory result has been achieved.

The following is one of the RGB filters I created:

Code 1. Matlab code of the RGB filter, song1

```
imgGray=B(:,:,1)>90 & B(:,:,1)<180 & B(:,:,2)>40 & B(:,:,2)<65 & B(:,:,3)>10 & B(:,:,3)<55;
```

To be precise, two new pairs of images have been obtained from each couple of frames of the original video. Those two images contain the hands of the pianist only.

Figure 8. Image taken from the side (RGB filter applied), song1

Figure 9. Image taken from above (RGB filter applied), song1

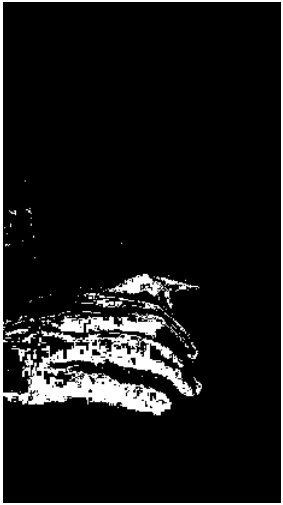

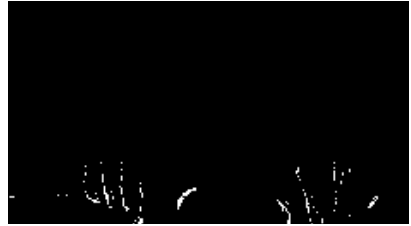

Figure 10. Image taken from the side (RGB filter applied), song2

Figure 11. Image taken from above (RGB filter applied), song2

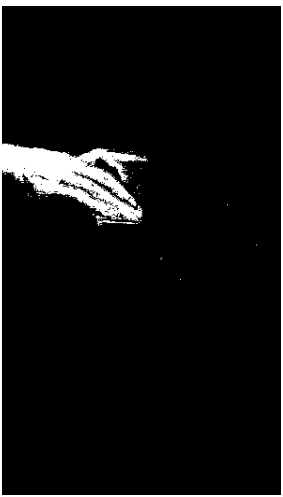

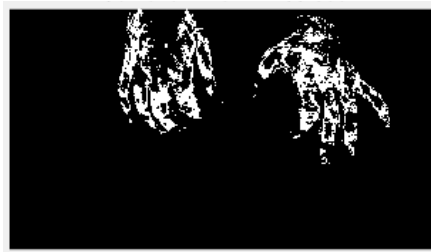

I noted that the extraction of the finger motion in the second scene was much more precise thanks to the homogeneous background dominated by a strong colour (blue in that case).

From that point, I proceed using the two videos whose recording was taken with the blue background in order to extract the motion of the fingers with higher accuracy.

After having removed some of the components of the binary image of the upper view of the hands having fewer pixels connected than a threshold (this operation was necessary because it removed many outliers that were left from the RGB filter), I created two images: one for each hand and I sized them.

Figure 12. Image taken from above (right hand extracted), song2    Figure 13. Image taken from above (left hand extracted), song2

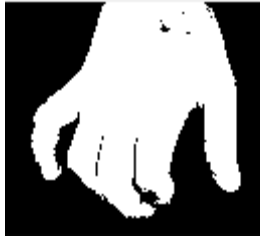    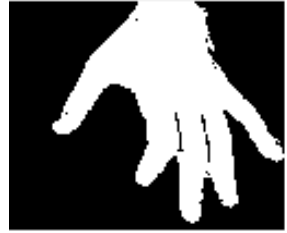

After that, I resized the images to try to centre the palm of the hands and extracted the edges of each of the images using the Canny method.

Figure 14. Image taken from above (estimation centre of the palm, left hand), song2

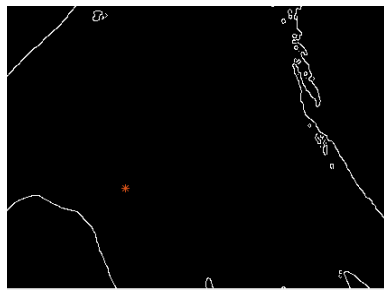

After that, as suggested in [2], I computed the distance transform from that image. I set to 0 the elements whose distance was lower than a threshold. The distance transform is the distance between a pixel and the closest pixel of the neighbourhood (that in that case is composed of the profile of the hand). The point having the maximum distance is the centre of the palm.

I computed the average of the coordinates of the points having as the value of distance the maximum value, so I obtained the coordinates of the centre of the palm.

Figure 15. Image taken from above (estimation centre of the palm, left hand), song2

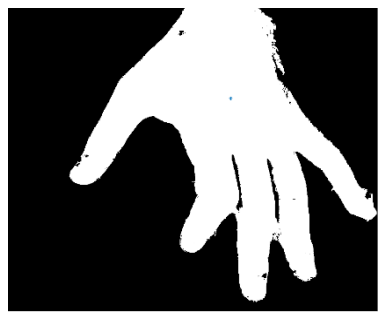

After that, I created a binary circle centred in the pixel coordinates of the centre of the palm and removed it from the images of the hands. The result consists of a pair of images, one for each hand, having the fingers extracted.

Figure 16. Image taken from above (extraction fingers right hand), song2

Figure 17. Image taken from above (extraction fingers left hand), song2

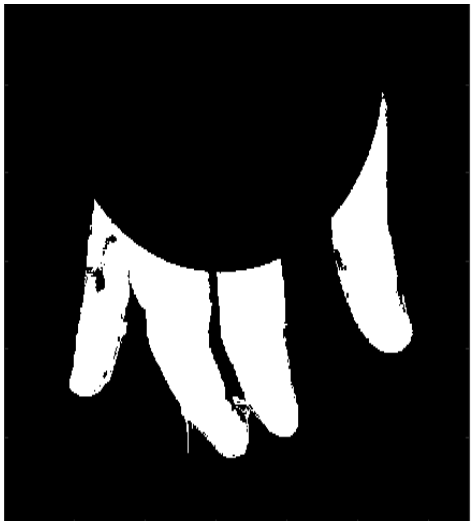
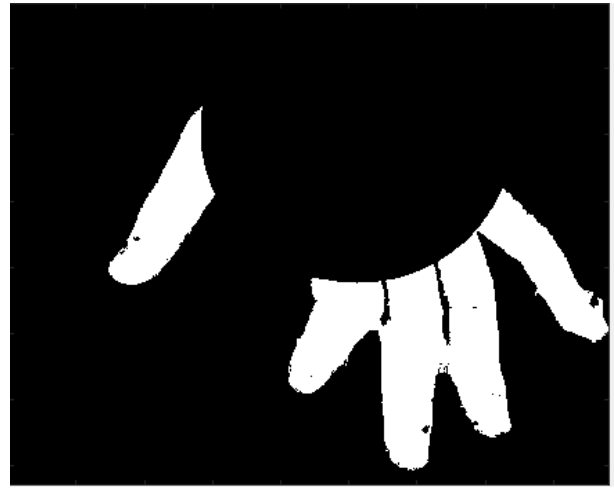

From that point, I identified the connected components of the binary images of the right and left hand. The connected components, in theory, should correspond perfectly to the fingers but in practice sometimes the fingers were too close to each other so they were recognized as part of the same connected component or they did overlap, resulting in being identified as part of the same connected component as well. For this reason, I was able to identify successfully most but not all the fingers of the hands.

Figure 18. Images taken from above (extraction fingers left and right hand), song2

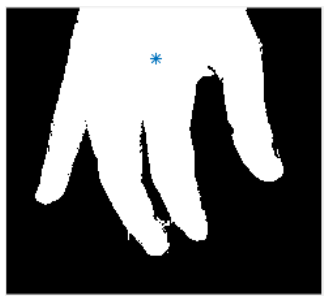
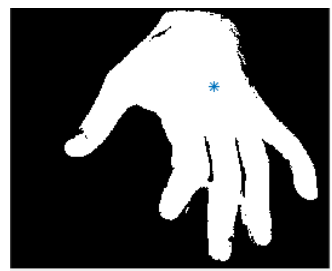
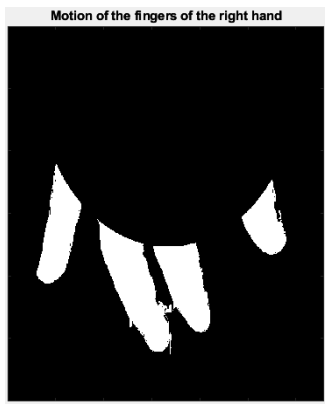
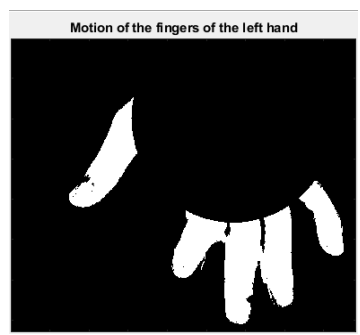

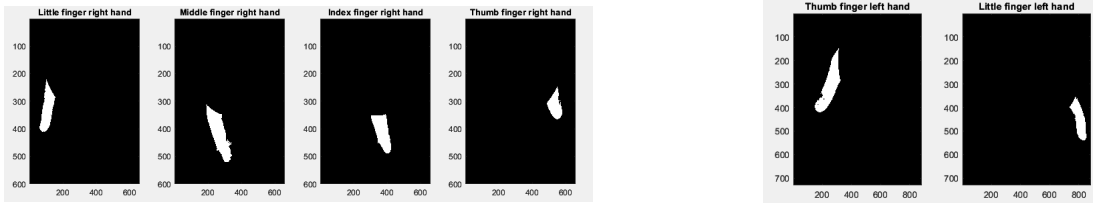

After that, I cropped the image of every single finger and rotated it also when necessary (I had to do so in the case of the middle and index fingers of the right hand).

Figure 19. Images taken from above (extraction fingers left and right hand), song2

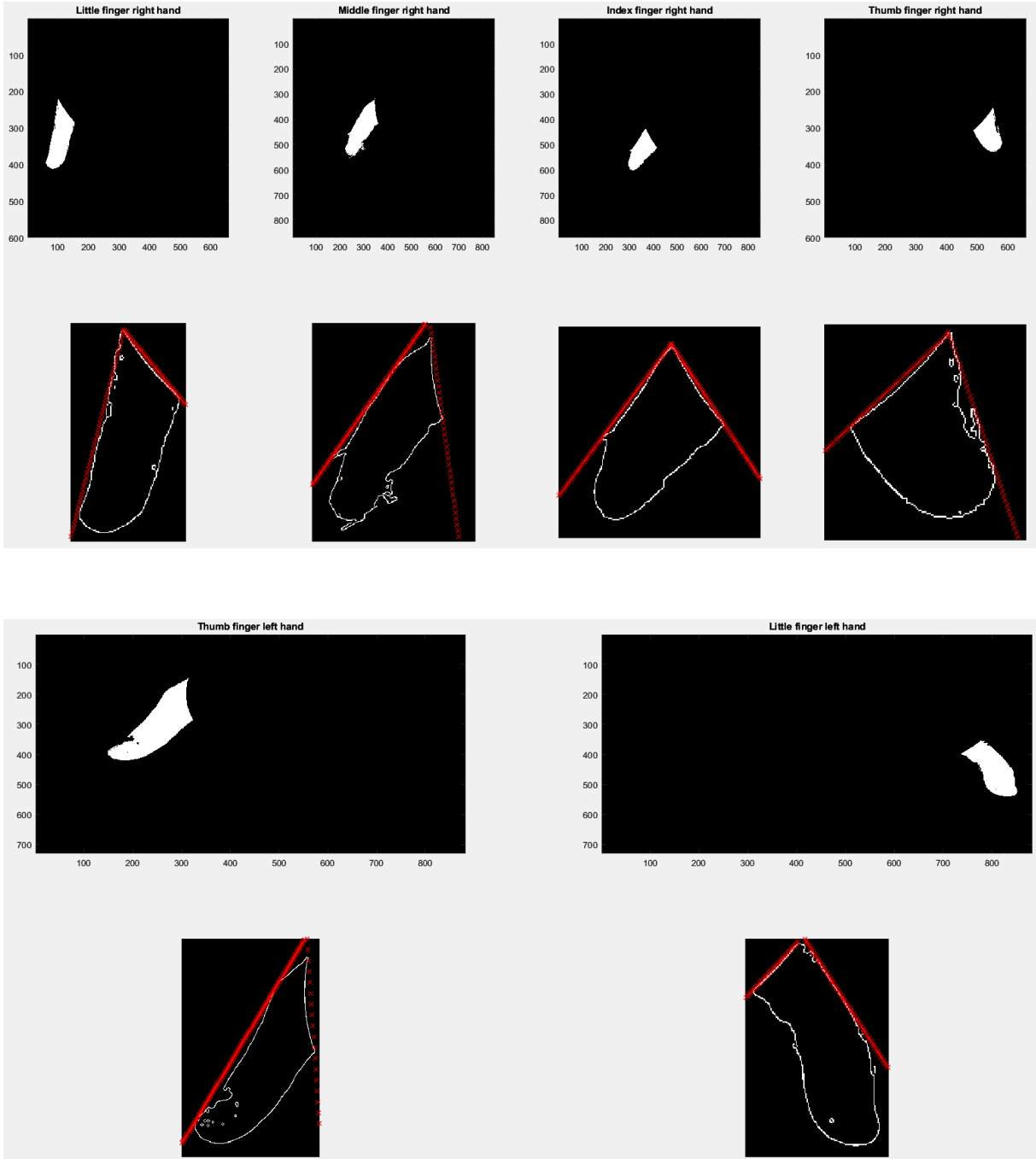

After that, I computed the perimeter of the convex hull of the images containing only the edges of the single fingers and computed a line that fitted the edges of the fingers from top-left to down-right and from top-right to down-left. Finally, I did draw some of the points of the lines.

From the image of the fingers with the two lines, it is quite easy to identify the three articulations because the edges move far away from the line as soon as a junction is met and also by observing the change of the inclination of the edges of the profile of the lower part of the finger.

Consider, for example, the little finger of the left hand.

Figure 20. Image taken from above (extraction little finger, left hand), song2

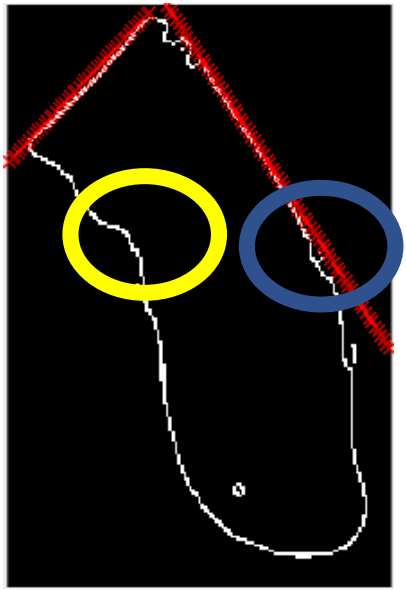

As clearly shown in the image of the extracted finger, a part of the profile of the finger match almost perfectly the line. In this case, as shown in the next image, the point from which the edges of the finger begin to differ from the line (that in this case is the area inside the blue circle), identify the last of the joints of the little finger (the one closest to the fingertip).

The same joint can also be identified by observing the change of the inclination of the edges of the profile of the lower part of the finger (that in this case is the area inside the yellow circle).

Figure 21. Identification of the joints of the fingers, right hand

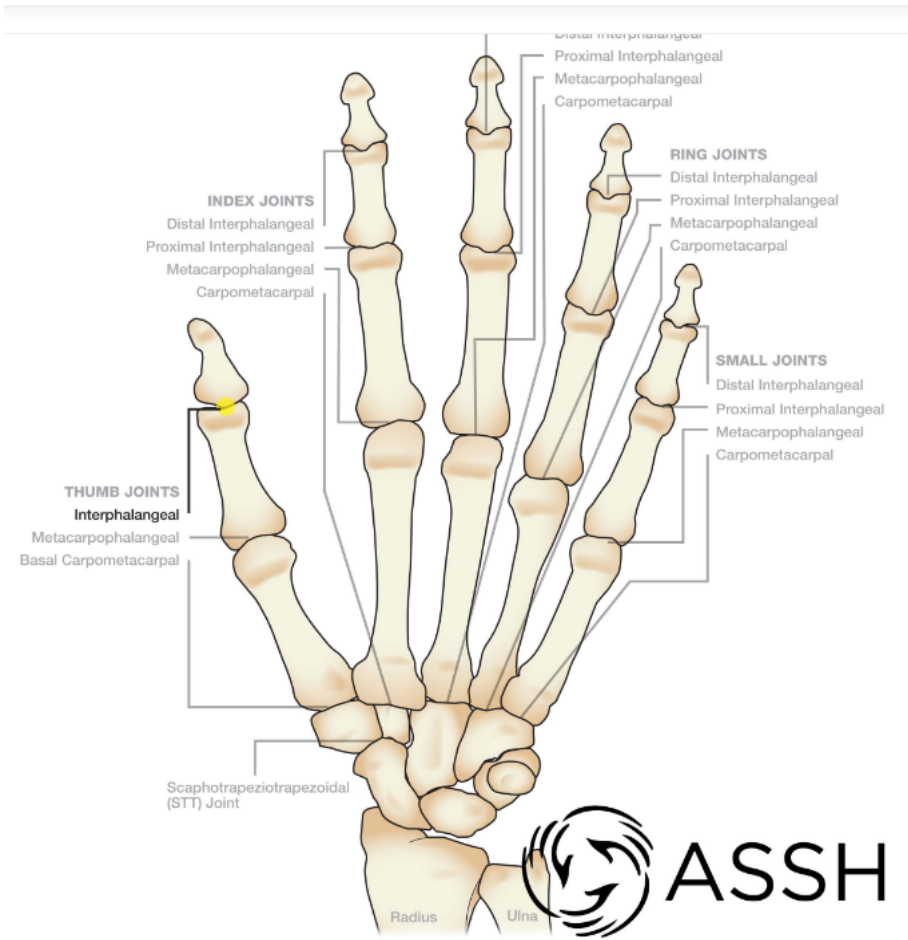

Image taken from: https://www.assh.org/handcare/safety/joints

Consider now the thumb finger of the left hand:

Figure 22.  Identification of the joints of the thumb finger, left hand

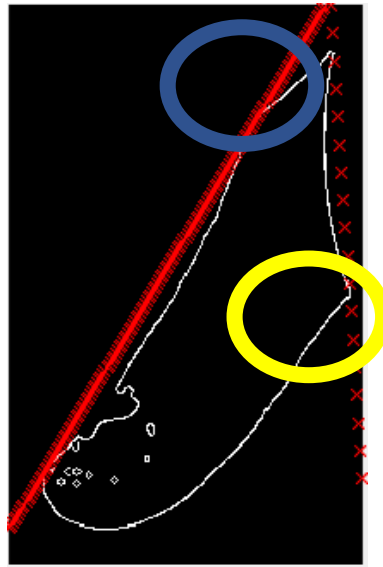

And also the little finger of the right hand:

Figure 23.  Identification of the joints of the little finger, right hand

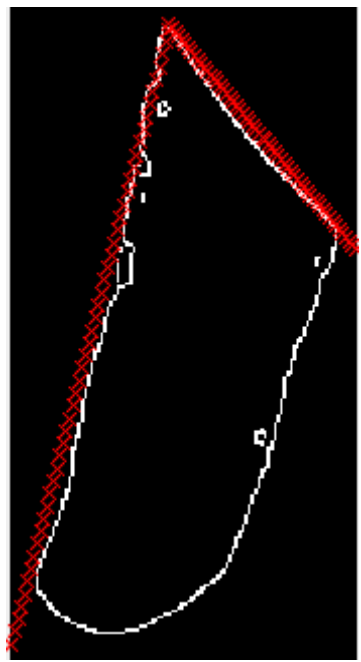

Consider now the side view:

Figure 24. Image taken from the side (RGB filter applied), song2

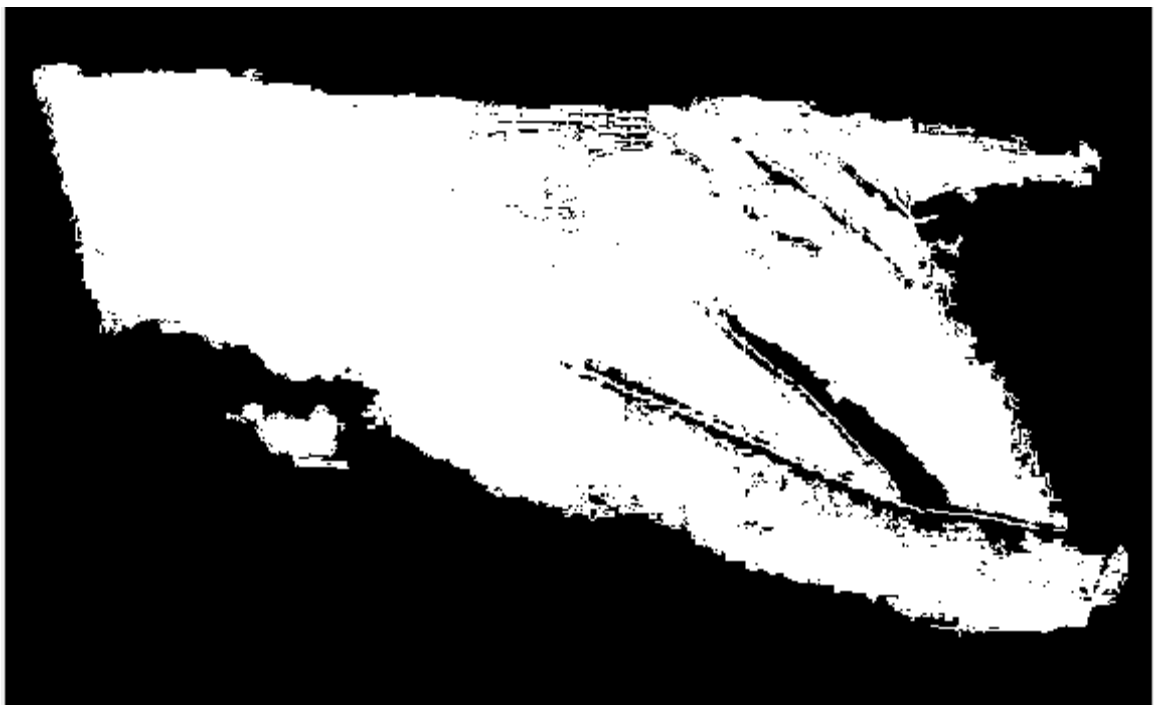

From that image I removed some little connected components and, similarly to what I did for the view from above, I computed the edges using the Canny method and computed the distance transform.

I computed the average of the coordinates of the points having as the value of the distance the maximum value, so I obtained the coordinates of a point that should identify a point close to the centre of both hands. Note that since the hands, from the side perspective, were very close, it was not possible to discriminate between the left and right hands.

Figure 25. Image taken from the side (distance transforms), song2

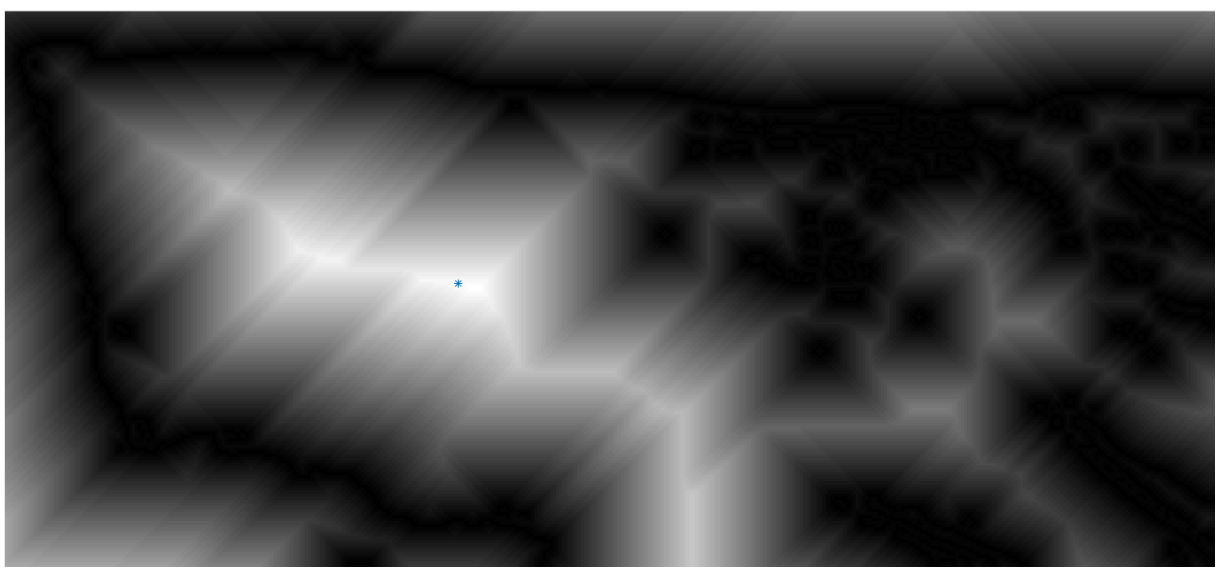

After having identified that point, I created a rectangle centred on it. After that, I rotated the binary image of the rectangle by an appropriate angle (32 degrees in this case) and subtracted it from the binary image of the hands taken from the side view.

The following is the result:

Figure 26.  Image taken from the side (fingers extracted), song2

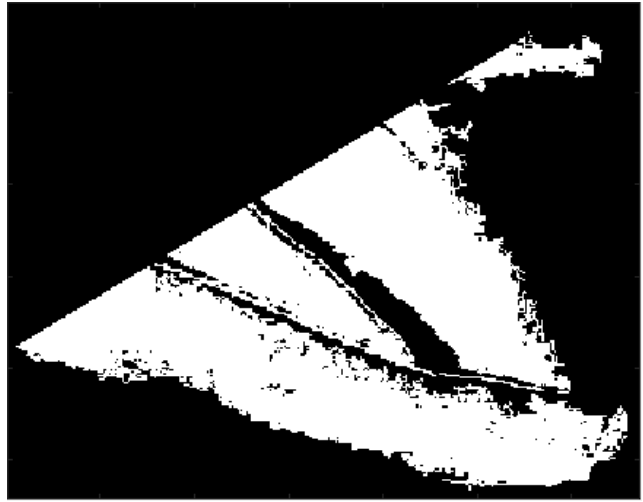

From this binary image, I proceeded with the extraction of the fingers by detecting the biggest connected components.

As seen during the extraction of the fingers from the upper view, sometimes the fingers are too close so they belong to the same connected component.

This is the case of the little and ring finger of the right hand as shown in the image:

Figure 27.  Image taken from the side (little and ring finger of the right hand extracted), song2

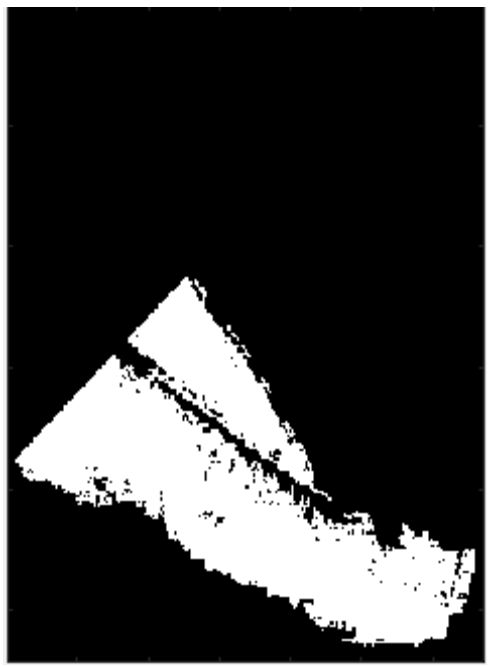

In order to discriminate the fingers, this time I decided to proceed manually by creating a suitable binary rectangle (rotated by an appropriate angle) and subtracting it from the image of both fingers.

The result is an image containing the little finger only.

Figure 28. Image taken from the side (little finger of the right hand extracted), song2

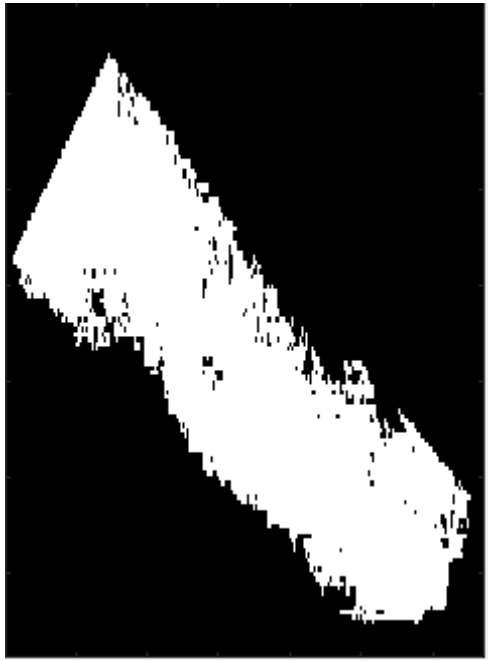

I obtained the image containing the ring finger only by subtracting the last image from the image of both the little and middle fingers of the right hand.

Figure 29. Image taken from the side (ring finger of the right hand extracted), song2

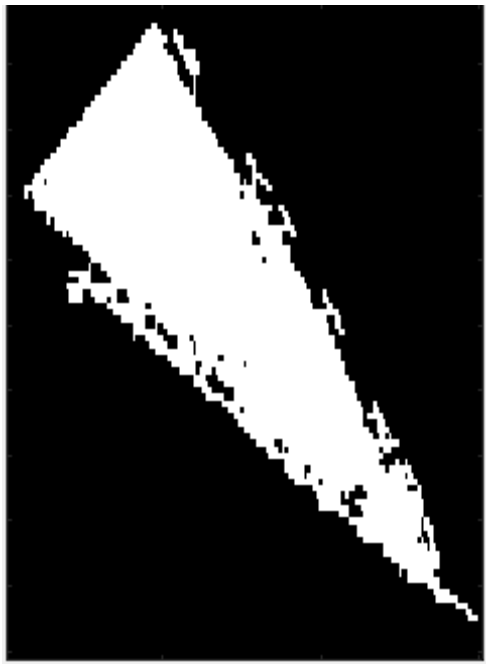

After that, I computed the perimeter of the convex hull of the images containing only the edges of the single fingers and computed a line that fitted the edges of the fingers from top-left to down-right and from top-right to down-left. Finally, I did draw some of the points of the lines.

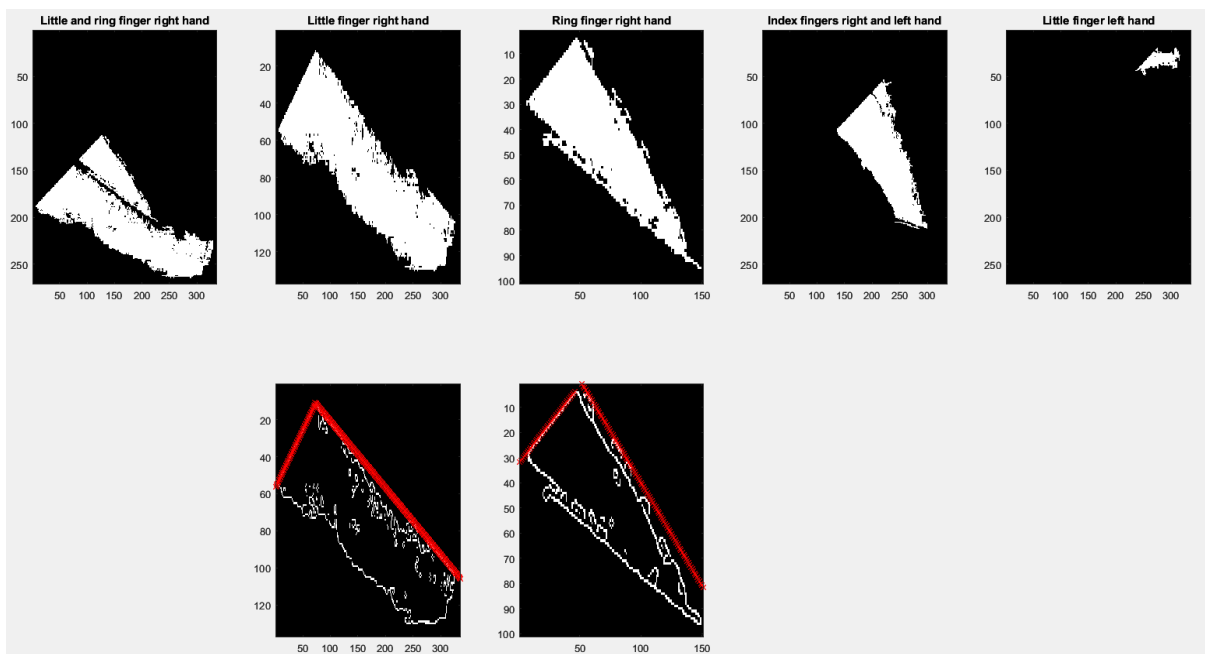
Figure 30. Image taken from the side (fingers extracted), song2

From the image of the fingers with the two lines it is quite easy to identify the three articulations because the edges move far away from the line as soon as a junction is met and furthermore, the junctions are easily identified by observing the change of the inclination in the edges of the profile of the lower part of the finger.

Consider, for example, the little finger of the right hand.

Figure 31. Image taken from the side (little finger of the right hand extracted), song2

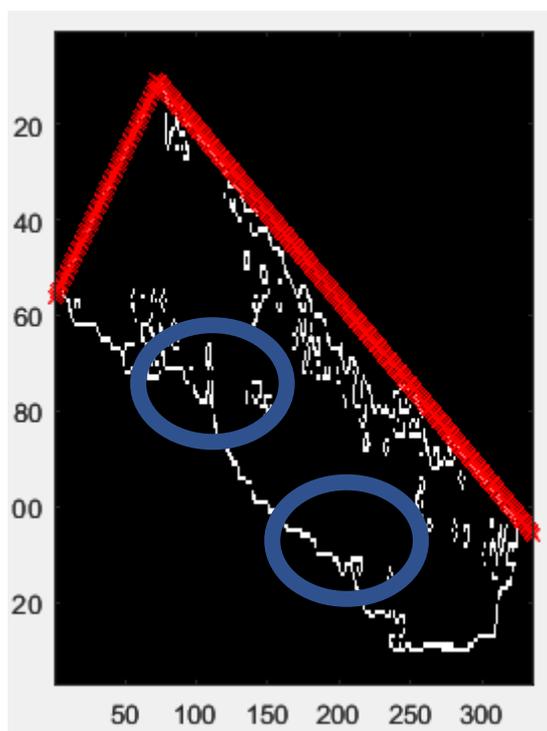

Two junctions can be easily detected by observing the change of the inclination in the edges of the profile of the lower part of the finger. In this case, the junctions are contained in the blue circles.

In this way, I was able to successfully extract the motion of the fingers of a piano player in their three articulations.

I consider the reached result quite satisfactory due to some little imprecisions in the computation of the centre of the palms (for what concerns the upper view) due to the fact that the images of the hands had to be cropped (such that the palm was roughly centred) in order to eliminate the values of the distances of the borders that were the highest but did not correspond to the centre of the palm. This fact implied that the coordinates of the centre of the palm had to be adjusted a little to compensate for this cropping. Another source of imprecision is due to the fact that I decided to adopt a fixed radius of the circle, differently from the approach used in [2], because this approach improved the precision of the detection of the correct radius, which otherwise would be probably underestimated due to the limited precision in the computation of the centre of the palm.

Regarding the work on the side view, some imprecisions in the computation of the point having the biggest distance transform values can be attributed to the noise that degraded the edges of the hands. Also, the fact that the hands were too close to each other and made it impossible to discriminate between the left and right hand degraded the quality of the result.

Another source of imprecision is the fact that I had to decide the angle of rotation of the rectangle that I subtracted from the image taken from the side in order to obtain the image containing the separated fingers.

Finally, the fact that I had to manually discriminate the little and ring finger of the right hand by creating an appropriate rectangle to be subtracted from the original image of the two fingers highlights the difficulty in distinguishing fingers that are close to each other and, as consequence, recognised as part of the same connected component.

The importance of using two videos taken from two different prospectives is highlighted when identifying the junctions of the fingers because without a side view it would be impossible to detect the junctions of the fingers in the case in which, using the upper view, the junctions are covered by the finger. One (or possibly two or more) side views can compensate for this defect, allowing more efficient and effective identification of the junctions even in those conditions.

Another note should be made about the fact that sometimes it was not possible to fully detect the three articulations of each finger. This is because of the imprecisions in the detection of the centre of the palm and also in the size of the radius of the circle, also to the fact that sometimes the fingers were too close to each other so they were recognized as part of the same connected component (despite my best efforts sometimes the noise corrupted the image so it was very hard to separate the image of the fingers that were close to each other) or they did overlap, resulting in being identified as part of the same connected component as well.

Another little detail is the fact that I had to analyse only the photograms in a certain temporal range in which the hands did not cross each other and the fingers of the two hands did not overlap or did not lie above the same key anyway.

A way to provide a better result could be to take other videos trying to record pieces in which the fingers of the pianist are not close to each other so that they are easy to discriminate during the processing phase and also in which the fingers of the two hands do not overlap and the hands do not cross each other, even if this would limit dramatically the input that it is possible to process by this algorithm.

A problem that is left open is a way to discriminate the left and right hand in the situation in which they cross each other (the discrimination of left and right hand is highly recommended to provide a better result when identifying the articulations of the fingers). Another problem is the fact that sometimes the fingers get too close to each other and, despite the recordings taken from two different points of view (from the above and from one side in the analysed case ), sometimes it is very hard to recognize them as different fingers.

**Identification of the pressed keys**

The starting point for the detection of the pressed keys is a video taken from the side of the piano and from above the keys level with an angulation of about 45 degrees wrt to the piano plane. This angulation will be useful to detect the pressed keys. The video capture all the region of the piano in which keys will be pressed.

This time, I made sure that the camera was not in contact with the keyboard and that both the camera and the keyboard were stable. Furthermore, I took care to illuminate the keyboard with a strong white light in order to remove most of the shadows that would degrade the quality of the result.

I started from the raw frames of the video, rotated them a bit and applied a dehaze filter.

This is the result:

Figure 32. Image taken from the side, song6

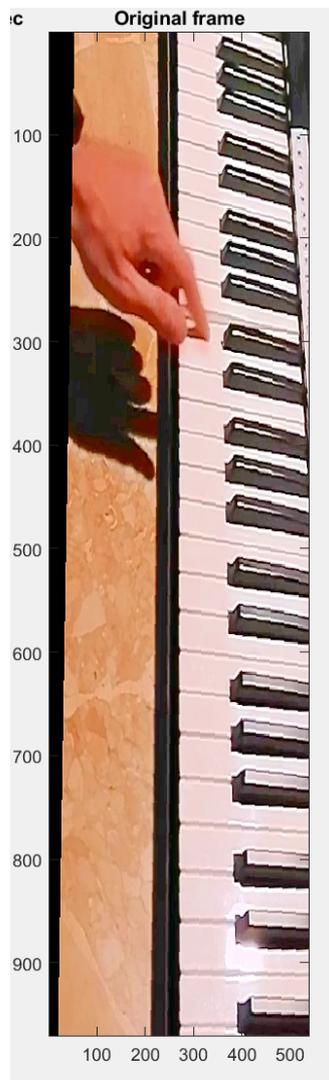

After that, I applied an RGB filter that I created to highlight the edges of the keys and cropped the width a bit if order to focus on the part of the keyboard that is bounded by the red rectangle in the following image.

Figure 33. Focus area of the keyboard

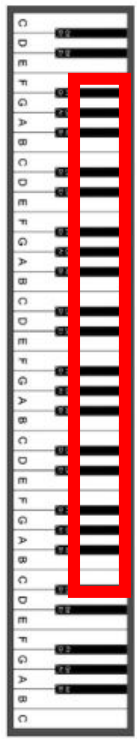

Image taken from https://playingkeys.com/61-keys-to-learn-piano/

The reason is that I noted that when a key is pressed, it is possible to observe a variation in the position of the pressed key in that area because of the position of the camera that is placed above the piano and its inclination wrt to the piano plane. Note that the hands of the pianist will not touch this part of the keyboard.

This is the result:

Figure 34. Identification key C3, song6

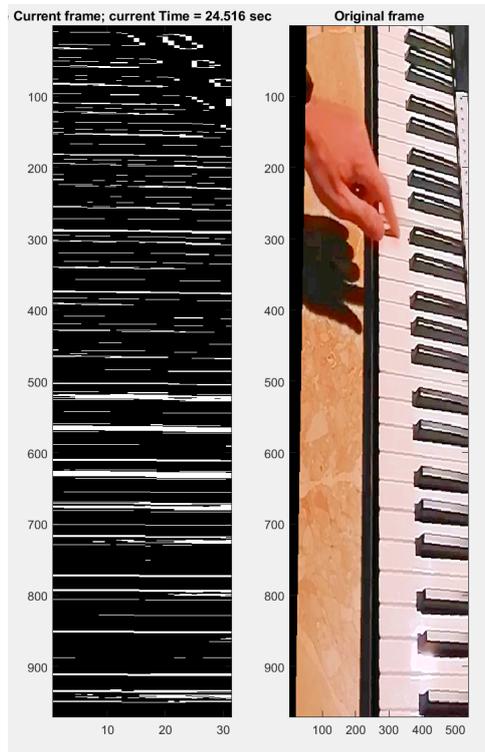

After that, I saved the last frame in which no keys are being pressed.

Figure 35. Image taken from the side (no key pressed), song6

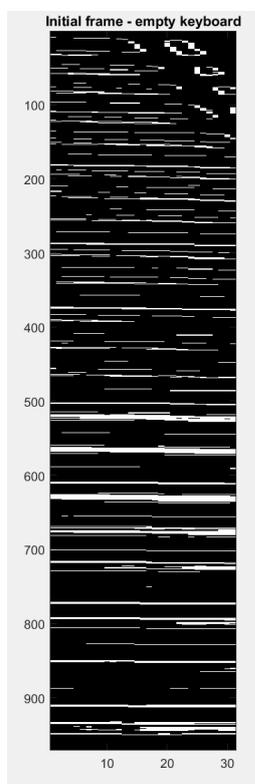

To detect the pressed keys, I will subtract that frame from the current frame as shown in the following image.

Figure 36. Difference of the images (key C3 pressed), song6

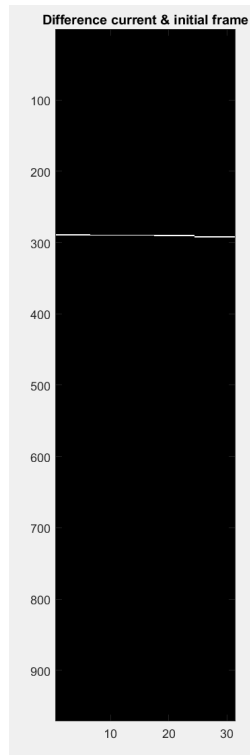

As clearly shown, after removing the smallest connected components, only the line (lines) corresponding to the left side (sides) of the key ( keys) that is (are) currently pressed is (are) left. This is because the camera was positioned on the right wrt to the piano.

After that, I created an RGB image starting from the raw current frame and setting to red the pixels corresponding to the pressed keys.

Figure 37. Identification key C3 pressed, song6

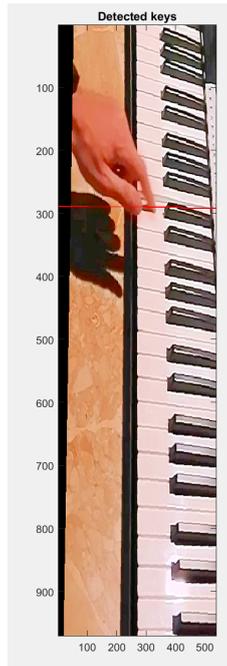

The name of the keys of the 61 keys piano (the one used in the video) is shown below:

Figure 38. Name of the keys in a 61 keys piano

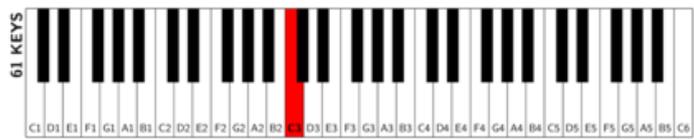

Source: https://i.stack.imgur.com/tQSxq.png

The following is the full result, and it shows the detection of the key C3:

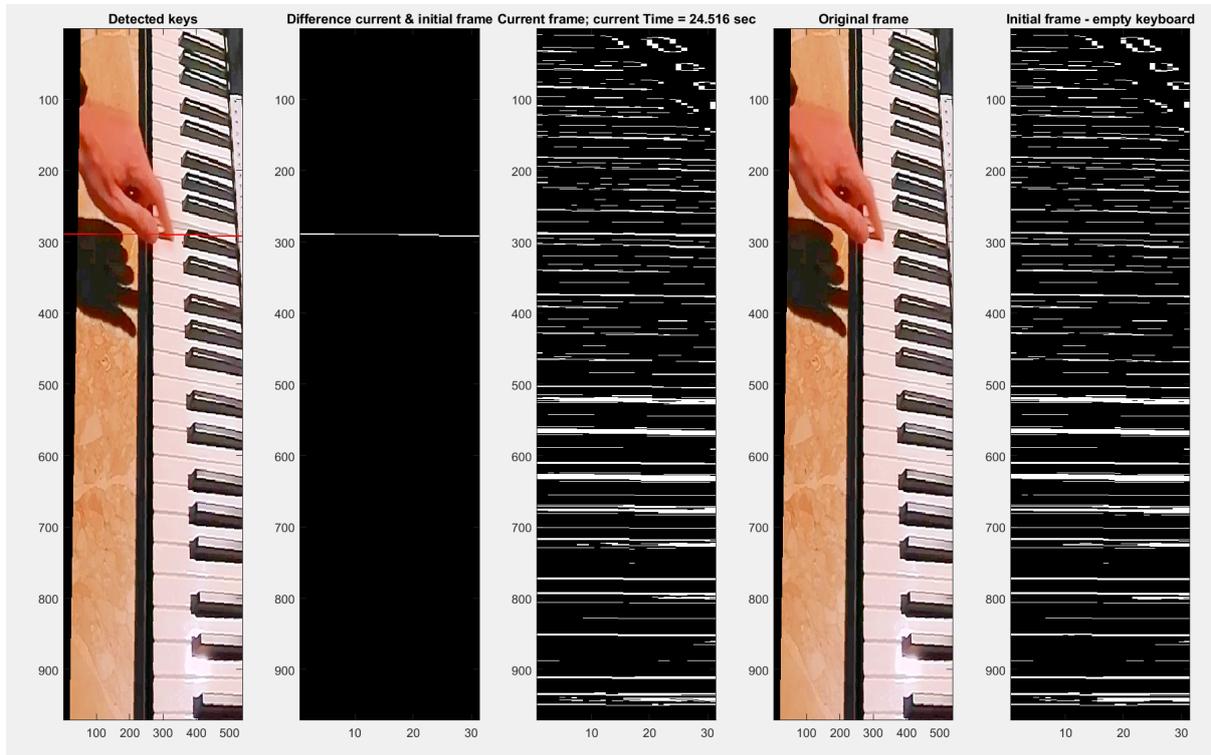

Figure 39. Identification key C3, song6

As clearly shown, the pressed key is easy to detect because its left side is highlighted in red. In this case, the pressed key is C3.

The following image shows the detection of the key C3b (D3#):

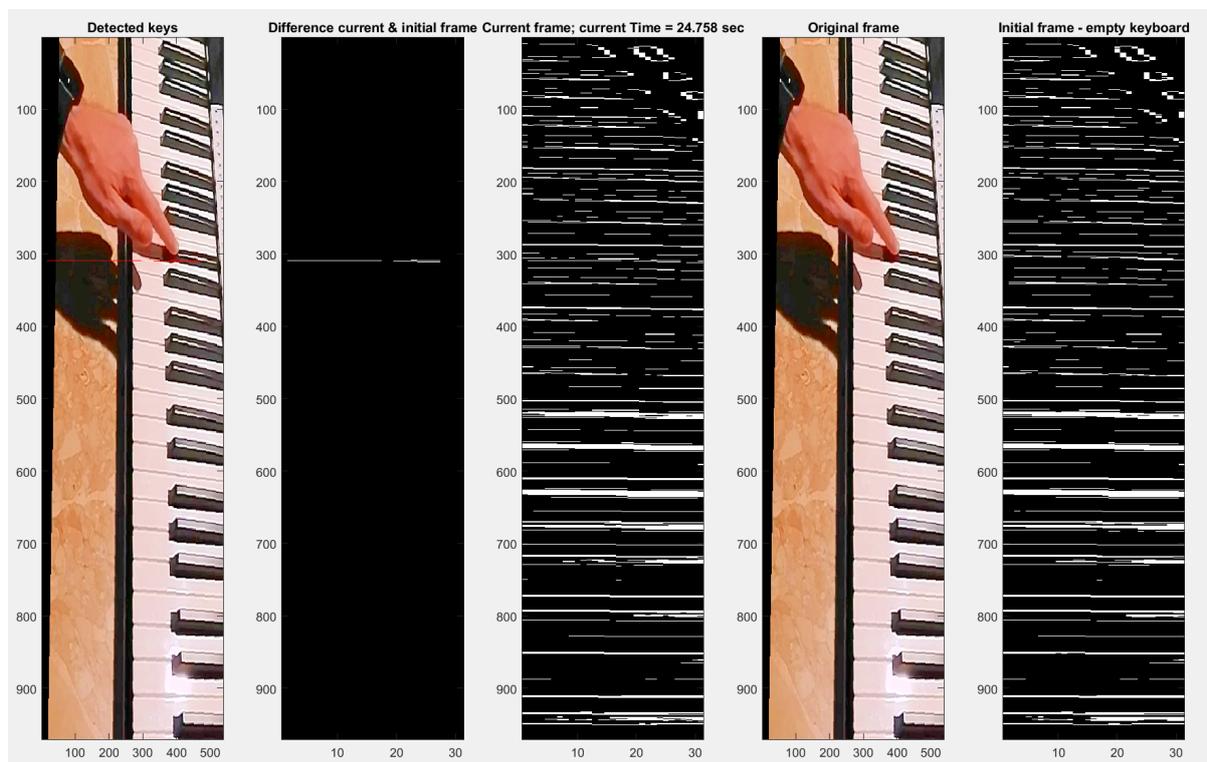

Figure 40. Identification key C3b, song6

The following image shows the detection of the key D3:

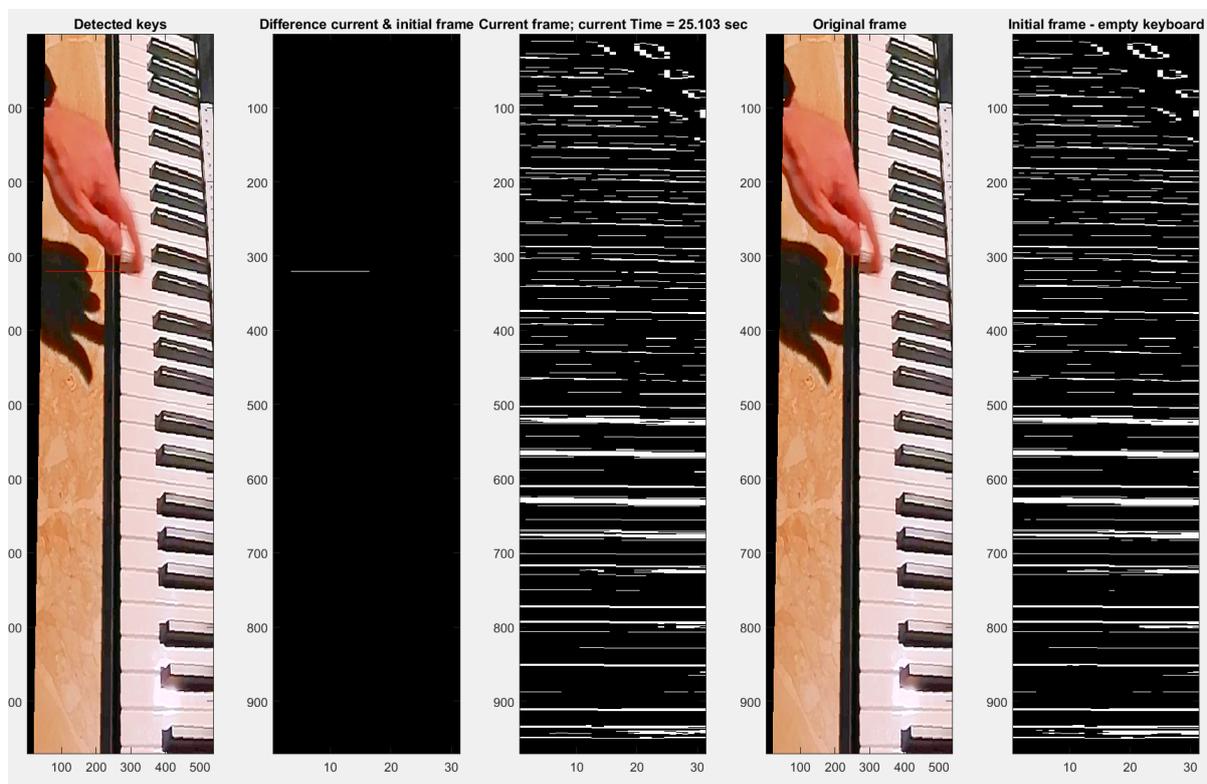

Figure 41. Identification key D3, song6

The following image shows the detection of the key D3b (E3#):

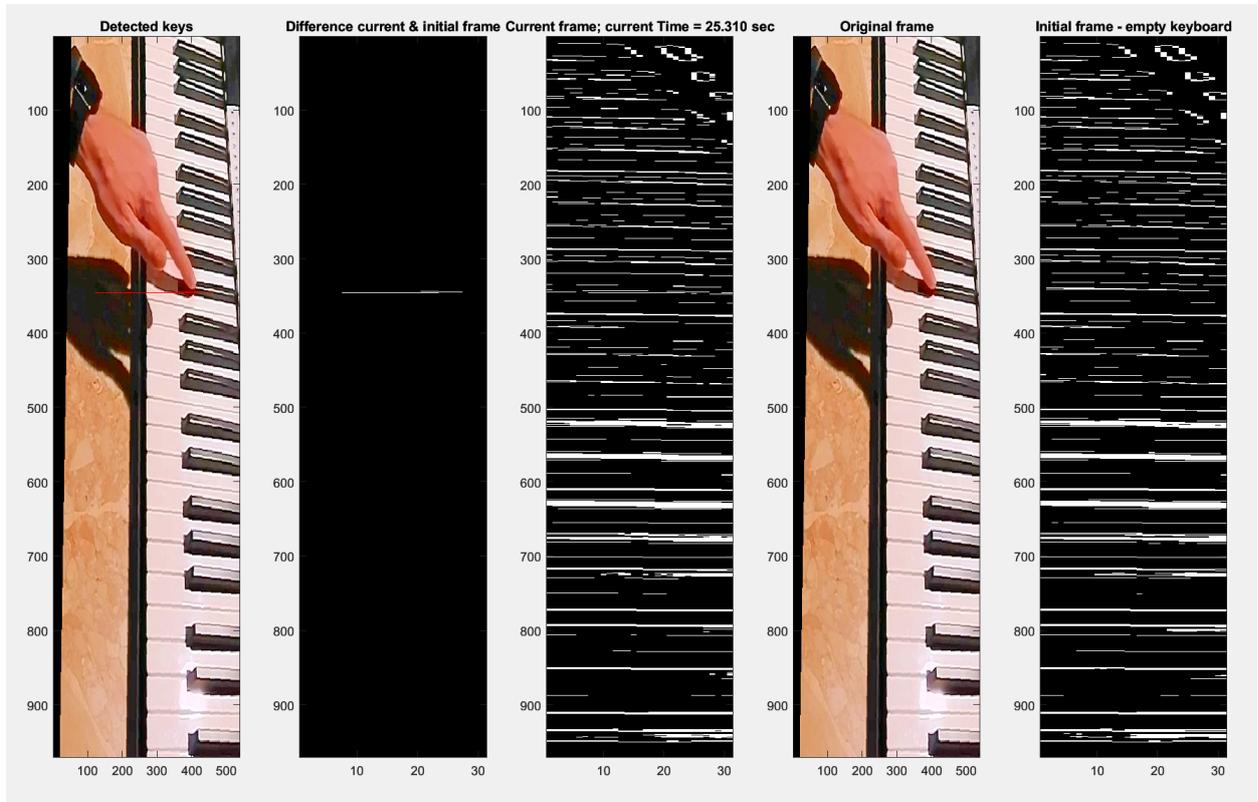

Figure 42. Identification key D3b, song6

The following image shows the detection of the keys C3, E3, G3, C4, E4, and G4 which are all pressed at the same time.

Figure 43. Identification keys C3, E3, G3, C4, E4, and G4

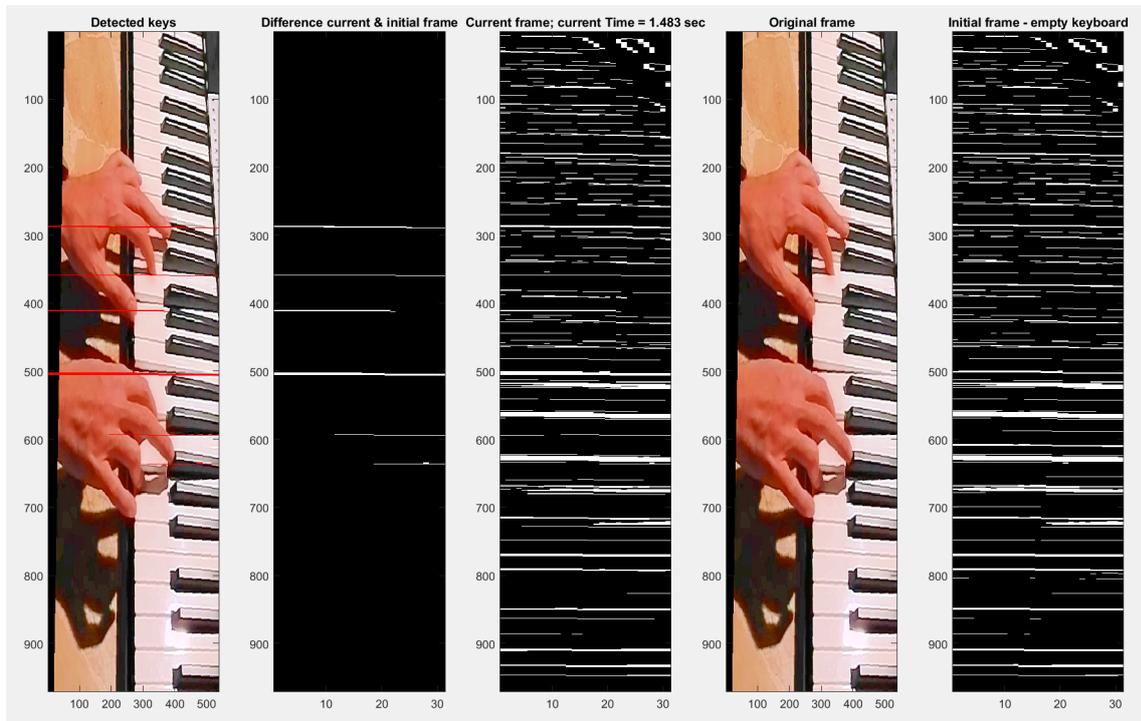

The following image shows the detection of the keys D3, F3b, A3, D4, F4b and A4 which are all pressed at the same time

Figure 44. Identification keys D3, F3b, A3, D4, F4b and A4

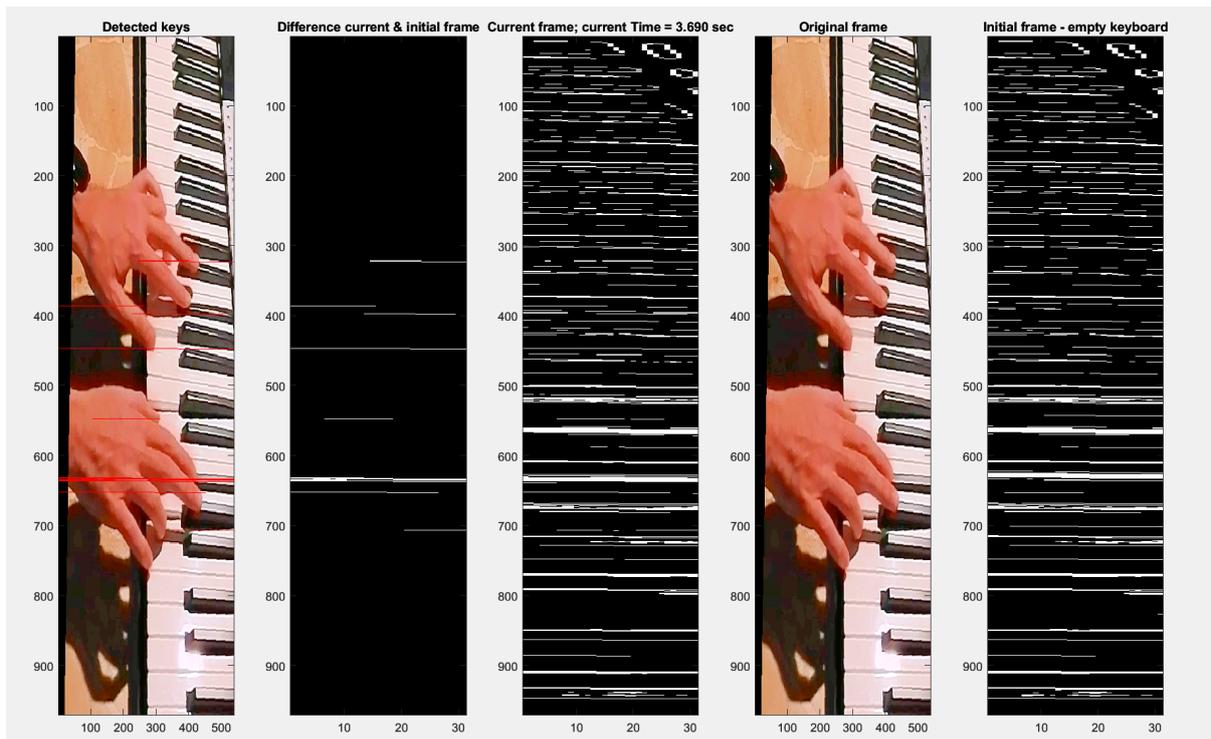

To identify the correct pressed keys I computed the top-left coordinates of each connected component that is in the difference frame and checked their value.

After having defined a suitable set of lower and upper bounds for the values of the coordinates for each key, I just checked for each frame whether the coordinates were part of one of the previously identified ranges that corresponded to a unique key.

Some examples of the ranges taken from the matlab code:

Code 2. Matlab code of the identification of the pressed keys, song6

```matlab
if(285<=upperCoordinateTile & upperCoordinateTile<=295)
   keyPressed= strcat(keyPressed, "C3; ");
end

if(305<=upperCoordinateTile & upperCoordinateTile<=315)
   keyPressed= strcat(keyPressed, "C3b; ");
end

if(320<=upperCoordinateTile & upperCoordinateTile<=330)
   keyPressed= strcat(keyPressed, "D3; ");
end

if(345<=upperCoordinateTile & upperCoordinateTile<=355)
   keyPressed= strcat(keyPressed, "D3b; ");
end
```

The following is the log of the matlab program that runs the example that was shown previously (keys C3, C3b, D3, and D3b are pressed in sequence ) in which the pressed keys are written together with the time.

Log 1. Matlab log of the identification of the pressed keys C3, C3b, D3, and D3b

ans =

   "Keys pressed: C3;  at 24.413 sec"

ans =

  "Keys pressed: C3;  at 24.482 sec"

ans =

  "Keys pressed: C3;  at 24.516 sec"

ans =

  "Keys pressed: C3b;  at 24.723 sec"

ans =

  "Keys pressed: C3b; C3b;  at 24.758 sec"

ans =

  "Keys pressed: C3b;  at 24.827 sec"

ans =

  "Keys pressed: D3;  at 25.103 sec"

ans =

  "Keys pressed: D3b;  at 25.275 sec"

As clearly shown by comparing the log with the previous images, the algorithm was able to successfully detect the pression of all 4 different keys.

The following is the log of the matlab program that shows the recognition of the keys E3, F3, F3b, G3, G3b, A3, A3b and B3 that were pressed after.

Log 2. Matlab log of the identification of the pressed keys E3, F3, F3b, G3, G3b, A3, A3b and B3

ans =

  "Keys pressed: E3;  at 25.551 sec"

ans =

  "Keys pressed: F3;  at 25.930 sec"

ans =

  "Keys pressed: F3;  at 25.965 sec"

ans =

    "Keys pressed: F3b;  at 26.206 sec"

ans =

    "Keys pressed: F3b;  at 26.309 sec"

ans =

    "Keys pressed: G3;  at 26.482 sec"

ans =

    "Keys pressed: G3;  at 26.516 sec"

ans =

    "Keys pressed: G3;  at 26.551 sec"

ans =

    "Keys pressed: G3b;  at 26.758 sec"

ans =

    "Keys pressed: A3;  at 27.034 sec"

ans =

    "Keys pressed: A3b;  at 27.378 sec"

ans =

    "Keys pressed: A3b;  at 27.413 sec"

ans =

    "Keys pressed: A3b;  at 27.482 sec"

ans =

    "Keys pressed: B3;  at 27.654 sec"

ans =

    "Keys pressed: B3;  at 27.827 sec"

The following is the log of the matlab program that runs the example that was shown previously (keys C3, E3, G3, C4, E4 and G4 are pressed all at the same time ) in which the pressed keys are written together with the time.

Log 3. Matlab log of the identification of the pressed keys C3, E3, G3, C4, E4 and G4

ans =

   "Keys pressed: E3; G3; C4; G4; E4;  at 0.414 sec"

ans =

   "Keys pressed: C4; E4; G4;  at 0.448 sec"

ans =

   "Keys pressed: C3; G3; C4; E3; E4; G4;  at 0.483 sec"

As clearly shown by comparing the log with the previous images, the algorithm was able to successfully detect the simultaneous pression of all 6 different keys.

The following is the log of the matlab program that runs the example that was shown previously (keys D3, F3b, A3, D4, F4b and A4 are pressed all at the same time ) in which the pressed keys are written together with the time.

Log 4. Matlab log of the identification of the pressed keys D3, F3b, A3, D4, F4b and A4

ans =

   "Keys pressed: A4;  at 3.586 sec"

ans =

   "Keys pressed: A4;  at 3.586 sec"

ans =

 "Keys pressed: D4; A4; F4b;  at 3.621 sec"

ans =

"Keys pressed: D4; A4; F4b;  at 3.621 sec"

ans =

  "Keys pressed: A3; D4; F4b; A4; F3b; A4; A3;  at 3.655 sec"

ans =

  "Keys pressed: F3b; A3; F4b; D4; D3; A4;  at 3.690 sec"

ans =

  "Keys pressed: F3b; A3; F4b; D4; D3; A4;  at 3.690 sec"

As clearly shown by comparing the log with the previous images, the algorithm was able to successfully detect the simultaneous pression of all 6 different keys.

Both white and black keys were identified correctly since the red line crosses the left side of the key that is being pressed, so in this way, it is possible to univocally discriminate between pressed white and black keys.

This is also shown in the following images in which the keys C3, C3b and D3 are being pressed:

Figure 45.  Identification key C3

C3

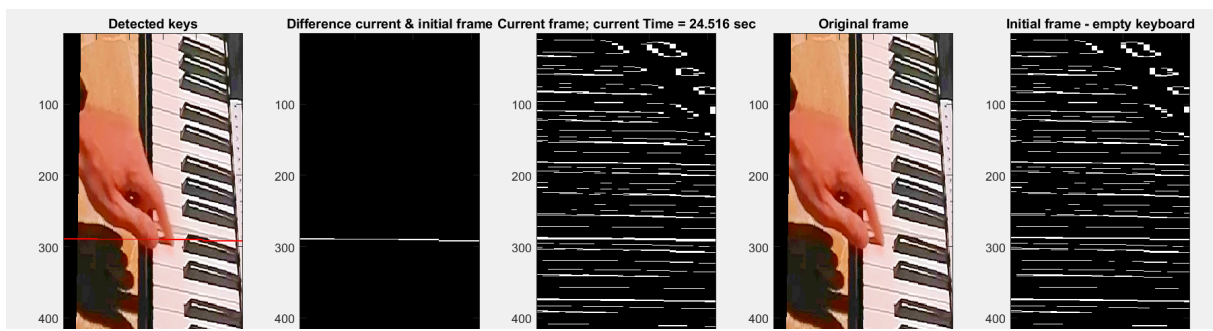

Figure 46.  Identification key C3b

C3b

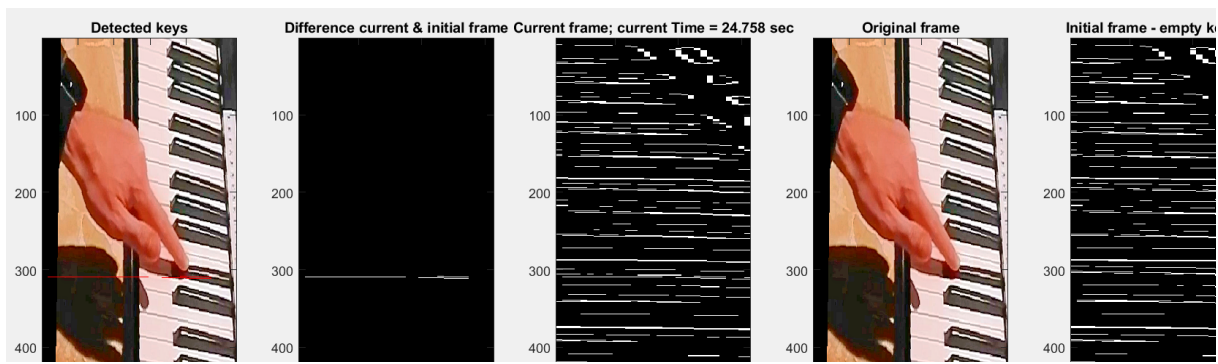

Figure 47. Identification key D3

D3

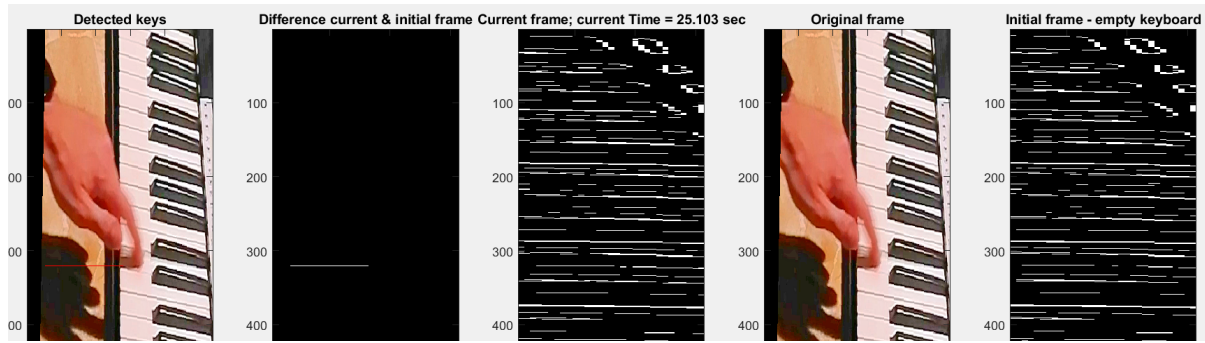

I consider the reached result quite satisfactory because I created an algorithm that is able to identify correctly the pression (even simultaneous) of piano keys.

A note should be made about the algorithm: it is crucial to create a suitable RGB filer to filter only the relevant part of the keyboard (almost no changes in the transformed frame should be detected when no keys are being pressed), otherwise, some false positives (incorrectly detected keys) may occur.

It is also very important to carefully define the ranges, aiming to reach a very high accuracy: this is because, if ranges are not set properly, the repeated pressing of the same key could be recognised as the pression of different keys.

Sometimes it takes some time to detect a pressed key (i.e. a key is not detected as pressed for some time before being recognized). This behaviour is probably caused by the slight change of the light (despite I did fix a lamp directly above the keyboard trying to aim to get the best possible result) and by the micro-vibrations that are caused by the pressure of the keys (despite I did collocate the piano on very stable support). The previously described causes of the problem could also be the reason because sometimes soft pressures of the keys are not detected (possibly also because of the smaller difference between the two frames or the too high threshold used to remove noisy connected components).

The choice of the threshold used to remove noisy connected components from the difference of the frames should be such that when no keys are pressed, the image should be black and that when one or more keys are pressed, the image should contain the connected components associated with the pressed keys only.

An idea to extend the already functional algorithm to create an automatic note transcriber could be to calibrate the system by showing the keyboard with no key pressed and then by pressing every single key (possibly many times in case high precision is needed) of the piano to automatically detect the ranges of the top y-coordinates of the connected components corresponding to the keys.

After that, by computing the duration of the pressure of the keys it is possible to compute the type of the piano note corresponding to the pressed key (whole, half-note, quarter-note, eight-note, etc…) and to create an output text (for example a pdf) containing the sheet music of the music that is being played.

Another similar application could be to create a system to automatically evaluate the performance of a pianist that could work as the previously described system with the difference that it would compare the played notes with the reference ones given as input (the one in the sheet music of the song that is being played).

A note should be made about the devices I used to create the examples and to test my approaches.

I used several devices to acquire the videos: two smartphones (a Huawei P30 lite having a camera with 48MP (f/1.8) Wide Angle lens, an 8MP Ultra-Wide Angle lens (f/2.4) and an 2MP dedicated bokeh lens and a Huawei Mate 9 having a camera with 8 MP, f/1.9, 26mm (wide), 1/3.2", 1.4µm, AF ) and one iPad having a camera with 8MP, ƒ/2.4 aperture.

The position of the cameras in the examples is the following:

**Extraction of the motion of fingers**

Figure 48. Position of the cameras in the extraction of the motion of fingers

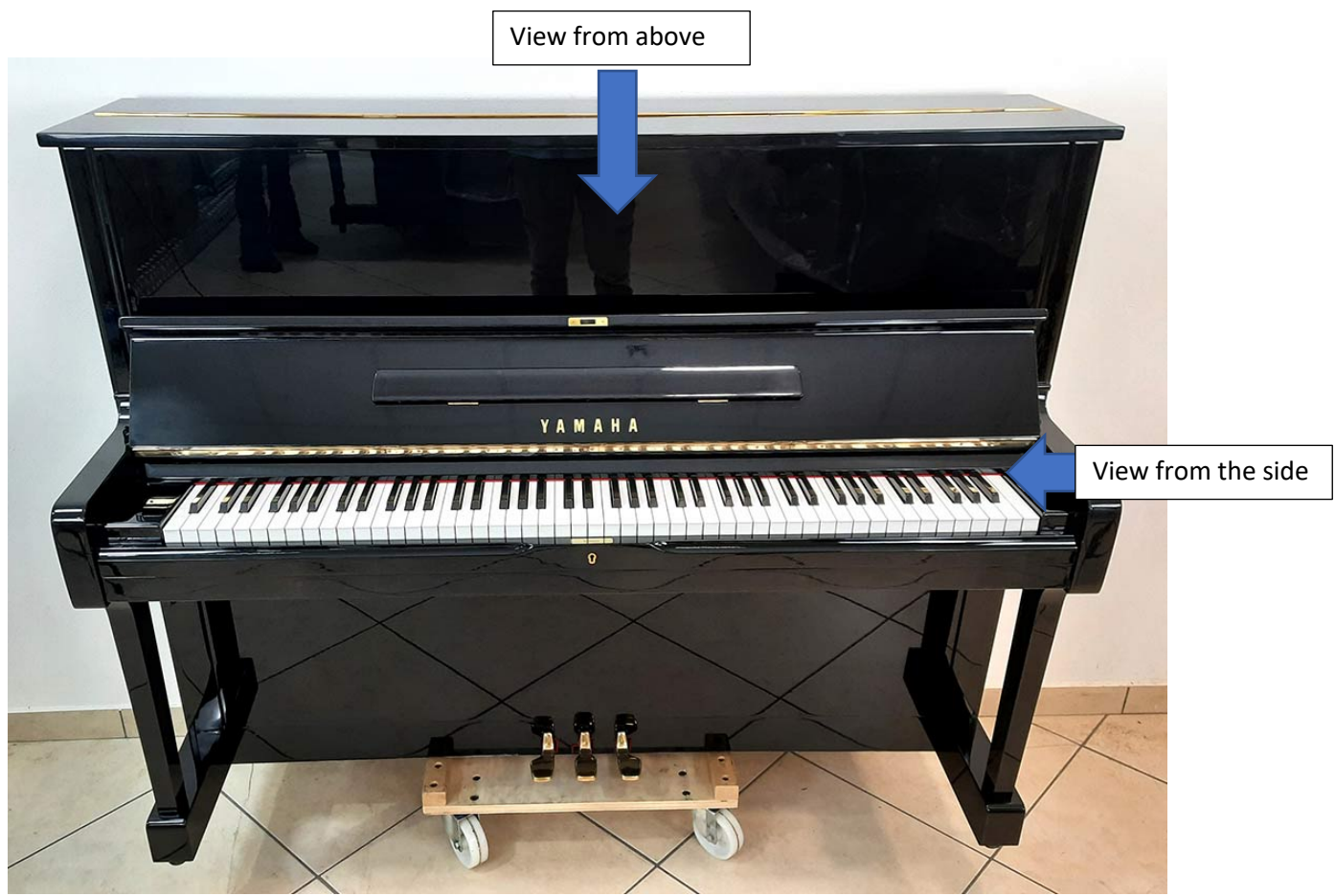

Source: https://pianofortiplaypiano.it/prodotto/pianoforte-verticale-yamaha-u1h/

**Identification of the pressed keys**

Figure 49. Position of the camera in the identification of the pressed keys

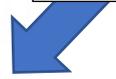
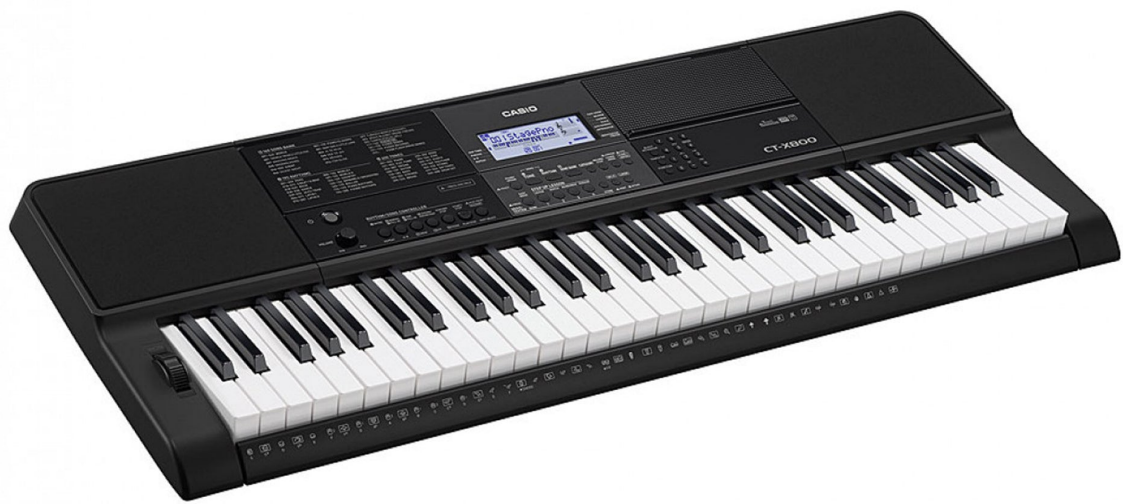

Source: https://crescendo.com.ph/collections/standard-keyboard/products/ct-x800-fa

In this work, I provided two approaches to solve two problems: the extraction of the motion of fingers (in their three articulations) of keyboard player and the identification of the pressed keys of a keyboard. I proceeded by analysing many related works and trying to replicate some of the techniques used, analysing the experimental results of the application of those approaches in practice (many times it was necessary to repeat the experiments several times due to some external factors, as the lighting, that cannot fully be controlled easily) and changing them when the results were too poor. I also used a mix of theoretical knowledge and intuition to find the cause of the problems that I faced during the process and to solve them. The results obtained were satisfactory and it is quite easy to modify my approaches to run different examples or to implement more complex functions.

# References

[1]  Temporal Control and Hand Movement Efficiency in
Skilled Music Performance - Werner Goebl, Caroline Palmer - January 3, 2013

[2] Real-Time Hand Gesture Recognition Using Finger Segmentation - Zhi-hua Chen, Jung-Tae Kim, Jianning Liang, Jing Zhang and Yu-Bo Yuan - 25 June 2014

[3] Pressed Piano Key Detection and Transcription by Visual Motion Analysis -
 Ali Özkaya, Sidem Işıl Tuncer – 2020-21

[4]  Detection of Piano Keys Pressed in Video - Potcharapol Suteparuk - Stanford University